\RequirePackage{fix-cm}
\documentclass[smallextended]{svjour3}       
\smartqed  
\usepackage{graphicx}
\usepackage{arabtex}
\usepackage[ruled]{algorithm2e} 
\usepackage{xcolor}
\usepackage{amsmath}
\usepackage{enumitem}
\usepackage{hhline}
\usepackage{url}
\usepackage{xcolor}
\usepackage{lineno,hyperref}
\modulolinenumbers[5]
\usepackage{arabtex}
\usepackage{utf8}
\usepackage[utf8x]{inputenc}
\usepackage[T1]{fontenc}
\usepackage{hhline}
\usepackage{url}
\usepackage{comment}
\usepackage{threeparttable}
\usepackage{xcolor}

\usepackage{booktabs} 
\usepackage{filecontents}

\begin{document}

\title{Sexism detection: The first corpus in Algerian dialect with a code-switching in Arabic/ French and English}
%



\author{Imane Guellil \and Ahsan Adeel\and Faical Azouaou\and Mohamed Boubred\and Yousra Houichi \and Akram Abdelhaq Moumna}


\institute{Imane Guellil \at
              Aston university; Birmingham, UK\\
              Folding Space, Birmingham, UK\\
              \email{i.guellil@aston.ac.uk}           
             \and
             Ahsan Adeel \at
             School of Mathematics and Computer Science, University of Wolverhampton, UK
              \and
           Faical Azouaou, Mohamed Boubred, Yousra Houichi, Akram Abdelhaq Moumna  \at
              Laboratoire des Méthodes de Conception des Systèmes. Ecole nationale Supérieure d’Informatique, 
BP 68M, 16309, Oued-Smar, Alger, Algérie. 
http://www.esi.dz
               \and 
               }


\maketitle

\begin{abstract}
In this paper, an approach for hate speech detection against women in Arabic community on social media (e.g. Youtube) is proposed. In the literature, similar works have been presented for other languages such as English. However, to the best of our knowledge, not much work has been conducted in the Arabic language. A new hate speech corpus (Arabic\_fr\_en) is developed using three different annotators. For corpus validation, three different machine learning algorithms are used, including deep Convolutional Neural Network (CNN), long short-term memory (LSTM) network and Bi-directional LSTM (Bi-LSTM) network. Simulation results demonstrate the best performance of the CNN model, which achieved F1-score up to 86\% for the unbalanced corpus as compared to LSTM and Bi-LSTM.
\end{abstract}

\section{Introduction}
In the literature, several hate speech definitions are adopted. However, the definition of Nockleby \cite{nockleby2000hate} has largely been used \cite{schmidt2017survey,zhang2018detecting,zhang2018hate}.  According to Nockleby, \textit{Hate speech is commonly defined as any communication that disparages or defames a person or a group based on some characteristic such as race, colour, ethnicity, gender, sexual orientation, nationality, religion, or other characteristic}. To illustrate how this hate could be presented in textual exchange, Schmidt et al. \cite{schmidt2017survey} presented a few examples such as 'Go fucking kill yourself and die already a useless, ugly pile of shit scumbag', 'Hope one of those bitches falls over and breaks her leg' etc.
Based on the recent survey of Schmidt et al. \cite{schmidt2017survey}, in this paper, the term \textit{Hate speech} is used as compared to \textit{abusive speech} which has widely been used in the literature \cite{andrusyak2018detection,gorrell2018twits}, \textit{offensive language} \cite{risch2018fine,pitsilis2018detecting,puiu2019semeval} or \textit{cyberbullying} \cite{dadvar2018cyberbullying,van2018automatic}. Chetty and Alathur \cite{chetty2018hate} categorised 
hate speech into four groups: gender hate speech (including any form of misogyny, sexism, etc.), religious hate speech (including any kind of religious discrimination, such as Islamic sects, anti-Christian, anti-Hinduism, etc.), Racist hate speech (including any sort of racial offence or tribalism, xenophobia, etc.) and disability (including any sort of offence to an individual suffering from health condition) \cite{al2019detection}. 

With the online proliferation of hate speech, a significant number of research studies have been presented in the last few years. The majority of these studies detect general hate speech \cite{burnap2014hate,davidson2017automated,wiegand2018inducing} and focused on detecting sexism and racism on social media \cite{waseem2016hateful,pitsilis2018detecting,kshirsagar2018predictive}. In contrast, only a few studies \cite{saha2018hateminers} focused on the detection of hate speech against women (only by distinguishing between hateful and non-hateful comments). However, almost all studies are dedicated to English where other languages such as Arabic is also one of four top used languages on the Internet\cite{guellil2018approche,guellil2018arabizi}). To bridge the gap, in this paper, we propose a novel approach to detect hate speech against women in the Arabic community. 

\section{Background}
\subsection{Hate speech}
The research literature adopts different definitions of hate speech. However, the definition of \cite{nockleby2000hate} was recently largely used by many authors such as,  \cite{de2018automatic,schmidt2017survey,zhang2018detecting,madisetty2018aggression} and \cite{zhang2018hate}.  According to Nockleby, \textit{"Hate speech is commonly defined as any communication that disparages or defames a person or a group based on some characteristic such as race, colour,
ethnicity,  gender,  sexual orientation,  nationality,
religion, or other characteristics"} \cite{nockleby2000hate}. For illustrating how this hate can be presented in textual exchange, \cite{schmidt2017survey} provided some examples:
\begin{itemize}
    \item Go fucking kill yourself and die already a useless, ugly pile of shit scumbag.
    \item The Jew Faggot Behind The Financial Collapse.
    \item Hope one of those bitches falls over and breaks her leg.
\end{itemize}
Based on the recent survey of \cite{schmidt2017survey}, we decided to use the term \textit{Hate speech} (which is the most commonly used) rather than other terms present in the literature for the same phenomenon such as: \textit{abusive speech} \cite{andrusyak2018detection,gorrell2018twits}, \textit{offensive language} \cite{risch2018fine,pitsilis2018detecting,puiu2019semeval} or \textit{cyberbullying} \cite{dadvar2018cyberbullying,van2018automatic}. According to Chetty and Alathur \cite{chetty2018hate},
hate speech is categorised into four categories: gendered hate speech, religious hate speech, racist hate speech and disability. Gendered hate speech includes any form of misogyny, sexism, etc. Religious hate speech includes any kind of religious discrimination, such as Islamic sects, Anti-Christian, anti-Hinduism, etc. Racist hate speech includes any sort of racial offence or tribalism, xenophobia, etc. Disability includes any sort of offence to an individual suffering from health which limits to do some of the life activities \cite{al2019detection}. 

\subsection{Arabic in social media}
Arabic is one of the six official languages of the United Nations\footnote{http://www.un.org/en/sections/about-un/official-languages/} \cite{eisele2010multiun,ziemski2016united,guellil2020arabic}. It is the official language of 22 countries. It is spoken by more than 400 million speakers. Arabic is also recognised as the 4th most used language of the Internet \cite{al2016prototype,boudad2017sentiment}. All the works in the literature \cite{habash2010introduction,farghaly2009arabic,harrat2017maghrebi,guellil2019arabic,guellil2020arautosenti} classify Arabic in three main varieties: 1) Classical Arabic (CA) which is the form  of Arabic language used in literary texts. The Quran \footnote{The Quran is a scripture which, according to Muslims, is 
the verbatim words of Allah containing over 77,000 words revealed through Archangel Gabriel to Prophet Muhammad over 23 years beginning in 610 CE. It is divided into 114  chapters of varying sizes, where each chapter is divided into verses, adding up to
a total of 6,243 verses. The work of Sharaf et al. \cite{sharaf2012qursim}} is  considered  to  be  the  highest  form  of CA text \cite{sharaf2012qurana}. 2) Modern Standard Arabic (MSA) which is used for writing as well as formal conversations. 3) Dialectal Arabic which is used in daily life communication, informal exchanges,etc \cite{boudad2017sentiment}. However, Arabic speakers on social media, discussion forums and Short Messaging System (SMS) often use a non standard romanisation called 'Arabizi' \cite{darwish2014arabizi,bies2014transliteration,guellil2020role}. For example, the Arabic sentence: \<rAny fr.hAnT>, which means I am happy, is written in Arabizi as 'rani fer7ana'. Hence, Arabizi is an Arabic text written using Latin characters, numerals and some punctuation \cite{darwish2014arabizi,guellil2018arabizi}. Moreover, most of Arabic people are bilingual, where the Mashreq side (Egypt, Gulf, etc) often use English and the Maghreb side (Tunisia, Algeria, etc) often use French, as second language. This linguistic richness contribute to increase a well known phenomenon on social media which is \textit{code switching}. Therefore, Arabic pages also contain messages such as: "\<rAny> super \<fr.hAnT>" or "\<rAny> very \<fr.hAnT>"  meaning I am very happy. In addition, messages purely written in French or in English are also possible.

Many work have been proposed, in order to deal with Arabic and Arabizi \cite{darwish2014arabizi,guellil2017arabizi}. Extracting opinions, analysing sentiments and emotion represent an emerging research area for Arabic and its dialects \cite{guellil2017approche,guellil2018sentialg,imane2019set}. However, few studies are dedicated to analyse extreme negative sentiment such as hate speech. Arabic hate speech detection is relatively a new research area where we were able to to collect only few works. These work are described in more details in the following section.
\section{Related work}
\subsection{Hate speech detection}

\subsubsection{General hate speech detection}
Burnap and Williams \cite{burnap2014hate} investigated the spread of hate speech after Lee Rigby murder in UK. The authors collected 450,000 tweets and randomly picked 2,000 tweets for the manual annotation conducted by CrowdFlower (CF) workers\footnote{https://www.figure-eight.com/}. Each tweet was annotated by 4 annotators. The final dataset contains 1,901 annotated tweets. The authors used three classification algorithms
and the best achieved classification results were up to 0.77 (for F1-score) using BLR. Davidson et al. \cite{davidson2017automated} distinguished between hateful and offensive speech by applying LR classifier 
The authors automatically extracted a set of tweets and manually annotated 24,802, randomly selected by CF workers. 
Their model achieved an F1 score of 0.90 but suffered poor generalisation capability with up to 40\% misclassification. 
Weigand et al. \cite{wiegand2018inducing} also focused on the detection of abusive language. The authors used several features and lexical resources to build an abusive lexicon. Afterwards, constructed lexicon in an SVM classification was used. In this work, publicly available datasets were used \cite{razavi2010offensive,warner2012detecting,waseem2016hateful}.

It is to be noted that all the aforementioned studies have been conducted with English language. However, a few other studies in some other languages are also conducted recently such as Italian \cite{del2017hate},
German \cite{koffer2018discussing}, 
Indonesian \cite{alfina2017hate}, Russian \cite{andrusyak2018detection}. However, only a limited number of researches have focused on hate speech detection in Arabic language. Abozinadah et al. \cite{abozinadah2015detection} evaluated different machine learning algorithms to detect abusive Arabic tweets. The authors manually selected and annotated 500 accounts associated to the abusive extracted tweets and used three classification algorithms 
The best results were obtained with the Naîve Bayes (NB) classifier with F1-score up to 0.90. Mubarek et al. \cite{mubarak2017abusive} focused on the detection and classification of the obscene and offensive Arabic tweets. The authors used the Log Odds Ration (LOR) 
For evaluation, the authors manually annotated 100 tweets and obtained a F1-score up to 0.60. Haidar et al. \cite{haidar2017multilingual} proposed a system to detect and stop cyberbullying on social media. The authors manually annotated a dataset of 35,273 tweets from Middle East Region (specially from Lebanon, Syria, Gulf Area and Egypt). For classification, the authors used SVM and NB and obtained the best results with SVM achieving F1-score up to 0.93. More recently, Alakrot et al. \cite{alakrot2018dataset} described a step by step construction of an offensive dataset of Youtube Arabic comments. The authors extracted 167,549 Youtube comments from 150 Youtube video. For annotation, 16,000 comments were randomly picked (annotated by 3 annotators). Finally, Albadi et al. \cite{albadi2018they} addressed the detection of Religious Arabic hate speech. The authors manually annotated 6,136 tweets (where 5,569 were used for training and 567 for testing). For feature extraction, AraVec \cite{soliman2017aravec} was used. Guellil et al.\cite{guellil2020detecting} also proposed a corpus for detecting hate speech against politician. This corpus is in Arabic/Algerian dialect. It includes 5,000 Youtube comments that were manually annotated by three annotators. For extracting features, the authors relied on both Word2vec and fastText. For classification both shallow and deep learning algorithms were used.

\subsubsection{Sexism detection (Hate speech against women)}
Waseem et al. \cite{waseem2016hateful} used LR classification algorithm to detect sexism and racism on social media. The authors manually annotated dataset containing 16,914 tweets where 3,383 tweets are for sexist content, 1,972 for racist content, and 11,559 for neither sexist or racism. For dataset generation, the authors used Twitter API for extracting tweets containing some keywords related to women. The authors achieved F1-score up to 0.73. The work of Waseem et al. \cite{waseem2016hateful} is considered as a benchmark by many researchers  \cite{al2019detection,pitsilis2018detecting,kshirsagar2018predictive}. The idea of Pitsilis et al. \cite{pitsilis2018detecting} is to employ a neural network solution composed of multiple Long-Short-Term-Memory (LSTM) based classifiers in order to detect sexism and racism in social media. The authors carried out many experiments achieving the best F1-score of 0.93. Kshirsagar et al. \cite{kshirsagar2018predictive} also focused on racism and sexism detection and their approach is also based on neural network. However, in this work, the author also used word embedding for extracting feature combining with a Muli-Layer Perception (MLP) based classifier. The best achieved F1-score was up to 0.71. Saha et al. \cite{saha2018hateminers} presented a model to detect hate speech against women. The authors used several algorithms to extract features such as bag-of-words (BOW), TF-IDF and sentence embeddings with different classification algorithms such as LR, XGBoost and CatBoost. The best achieved F1-score was 0.70 using LR classifier. Zhang et al. \cite{zhang2018detecting} proposed a hybrid model combining CNN and LSTM to detect hate speech. The authors applied their model on 7 datasets where 5 are publicly available \cite{waseem2016hateful,waseem2016you,gamback2017using,park2017one,davidson2017automated}.

\subsection{Motivation and contribution}
The hate speech detection on social media is relatively a new but an important topic. There are very few publicly available corpora mostly dedicated to English. Even for English, less than 10 resources are publicly available. More recently, researchers have presented  work in other languages including German, Italian, Arabic. However, most of the work focuses on detecting a general hate speech not against a specific community. On Arabic, only 5 research studies are presented in the literature which are mainly focused on Twitter. This paper focuses on Youtube which is the second biggest social media platform, after Facebook, with 1.8 billion users \cite{kallas2017top,alakrot2018dataset}. The major contributions of this study are: Development of a novel hate speech corpus against women containing MSA and Algerian dialect, written in Arabic, Arabizi, French, and English. The corpus constitutes 5,000 manually annotated comments. For corpus validation, three deep learning algorithms (CNN, LSTM, and bi-LSTM) are used for hate speech classification. For feature extraction, algorithms such as word2vec, FasText, etc., are used.


\section{Methodology}
\begin{figure} 
		\centering\includegraphics[scale=1.0,width=1.0\textwidth]{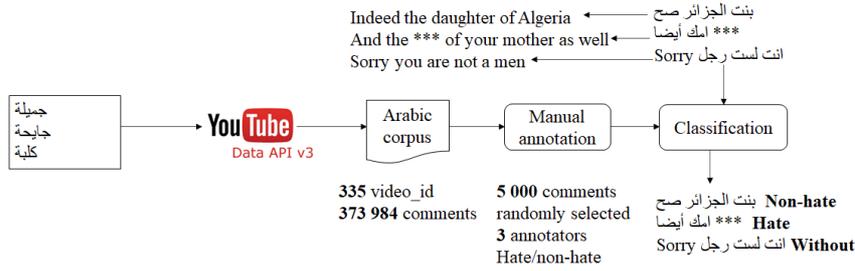}
		\caption{The general steps for constructing and validating the annotated corpus}
		\label{archi}
	\end{figure}

Figure \ref{archi} illustrates the general steps for constructing the annotated corpus for classifying hate speech regarding women. This figure includes some examples as well.
\subsection{Dataset creation}
\subsubsection{Data collection}
Youtube comments related to videos about women are used. Feminine adjective such as: \<jmylT> meaning beautiful, \<jAy.hT> meaning stupid or \<klbT> meaning a dog are targeted. A video on Youtube is recognised by a unique identifier (\textit{video\_id}). For example the video having an id equal to \textit{"TJ2WfhfbvZA"} handling a radio emission about unfaithful women and the video having an id equal to \textit{"\_VimCUVXwaQ"} gives advices to women for becoming beautiful. Three annotator, manually review the obtained video from the keyword and manually selected 335 \textit{video\_id}. We used Youtube Data API\footnote{https://developers.google.com/youtube/v3/} and a python script to automatically extract comments of each \textit{video\_id} and their replies. At the end, we were able to collect \textit{373,984} comments extracted for the period between February and March 2019, we call this corpus \textit{Corpus\_Youtube\_women}.

\subsubsection{Data annotation}
For the annotation, we randomly select 5,000 comments. 
The annotation was done by three annotators, natives speaker of Arabic and its dialects. The annotators was separated and they had one week for manually annotated the selected comments using two labels, 1 (for hate) and 0 (for non-hate). The following points illustrate the main aspects figuring in the annotators guideline:
\begin{itemize}
\item The annotators should classify each comments containing injuries, hate, abusive or vulgar or offensive language against women as a comment containing hate.
\item The annotators should be as objective as they can. Even if they approve the comment, they should consider it as containing hate speech is it is offensive against women.
\item For having a system dealing with all type of comments, the annotators were asked to annotate all the 5,000 comments, even if the comment speak about football or something not related to women at all. However they asked to annotate this comment with 0 and to add the label w (meaning without interest).
\item When the annotators are facing a situation where they really doubt about the right label, they were asked to put the label p (for problem) rather than putting a label with which they are not convinced.
\end{itemize}
At the beginning of the annotation process, we received lots questions such as: 1) Have the hate have to be addressed to women, how to classify a message containing hate regarding men? 2) Have the hate comments absolutely contains terms indicating hate or have the annotators to handle irony?, etc. For the first question, we precise that the comments have to be addressed to women. Any others comment have to be labelled with 0 
For the second question, we asked the annotators to also consider the irony and sarcasm. 

After completion of the annotation process, we concentrate on the comments obtaining the same labels from all annotators. Then, we constructed two dataset. The first one (\textit{Corpus\_1}) contains 3,798 comments which are annotated with the same labels (0 or 1) from the three annotators. Among this comments 792 (which represent 20.85\%) are annotated as hateful and 3006 as non-hateful. Hence, this corpus is very unbalanced. The second one (\textit{Corpus\_2}) represents the balanced version of (\textit{Corpus\_1}). For constructing this corpus, we randomly picked up 1,006 comments labelled as non-hateful and we picked up all the comments annotated as hateful. Then, we constructed a balanced corpus containing 1,798 comments.

\begin{figure} 
		\centering\includegraphics[scale=1.0,width=1.0\textwidth]{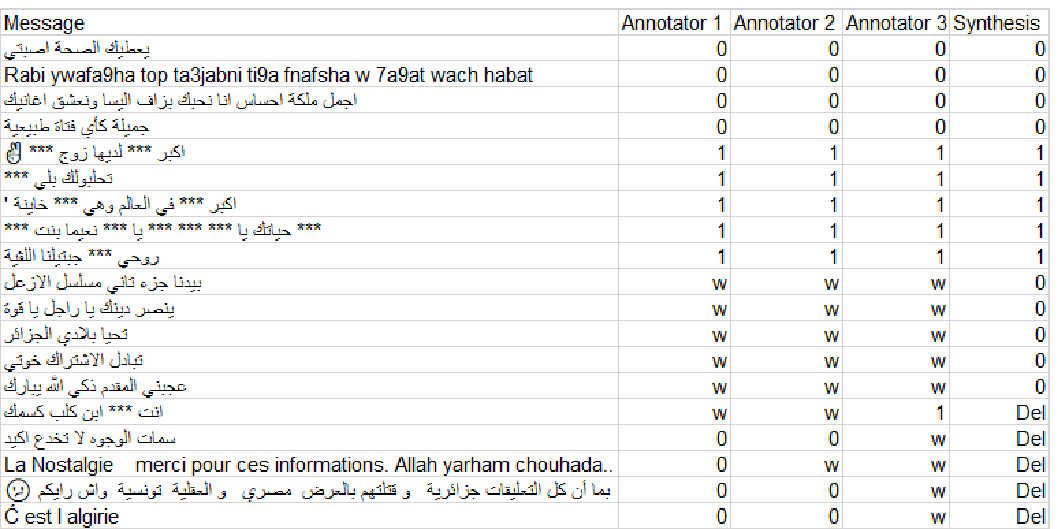}
		\caption{Sample of the annotated corpus}
		\label{sample}
	\end{figure}

For better illustrating the annotated data, we present Figure \ref{sample}. The column message contains some comments (extracted from Youtube) and manually annotated. The following three columns illustrate the annotation given by each one of the annotators. When all the annotators agree, the messages are kept in the corpus. In the other case, they are removed.  The removed messages are not considered for the training (we proceed in this manner in order to increase the precision. We also plan to extend this corpus in the future automatically, hence, the precision is crucial). If the message is annotated as hateful by all the annotators, it is then annotated as hateful and column hate receives 1. In the other case, it receives 0.

\subsection{Hate speech detection}
\subsubsection{Features extraction}
We use two different algorithm for features extraction which are, Word2vec \cite{mikolov2013efficient} and FasText \cite{joulin2016bag}. We use Word2vec with classic methods and we use FasText with Deep learning methods. Word2vec describes two architectures for computing continuous vectors representations, the Skip-Gram (SG) and Continuous Bag-Of-Words (CBOW). The former predicts the context-words from a given source word, while the latter does the inverse and predicts a word given its context window \cite{mikolov2013efficient}. As for Word2vec, Fastext models is also based on either the skip-gram (SG) or the continuous bag-of-words (CBOW) architectures. The key difference between FastText and Word2Vec is the use of n-grams. Word2Vec learns vectors only for complete words found in the training corpus. FastText learns vectors for the n-grams that are found within each word, as well as each complete word \cite{grave2018learning}. In this work we rely on both representations of word2vec and fasText (i.e SG and CBOW). 

For Word2vec model, we used the Gensim toolkit\footnote{https://radimrehurek.com/gensim/models/word2vec.html}. For fasText, we use the fasText library proposed by Facebook on Github\footnote{https://github.com/facebookresearch/fastText}. For both Word2vec/fasText, we use a context of 10 words to produce  representations for both CBOW and SG of length 300. We trained the Word2vec/fasText models on the corpus \textit{Corpus\_Youtube\_women}

\subsubsection{Classification}
For comparing the results, we use both classification methods, classic and deep learning based. For classic method, we use five classification Algorithms such as: GaussianNB (GNB), LogisticRegression (LR), RandomForset (RF), SGDClassifier (SGD, with loss='log' and penalty='l1') and LinearSVC (LSVC with C='1e1'). For their implementation phase, we were inspired by the classification algorithm proposed by Altowayan et al. \cite{altowayan2016word}. For the deep learning classification we use three models CNN, LSTM and Bi-LSTM. For each model, we use six layers. The first layer is a randomly-initialised word embedding layer that turns words in sentences into a feature map. The weights of embedding\_matrix are calculated using fasText (with both SG and CBOW implementation). This layer is followed by a CNN/ LSTM/BiLSTM layer that scans the feature map (depending on the model that we defined). These layers are used with 300 filters and a width of 7, which means that each filter is trained to detect a certain pattern in a 7-gram window of words. Global maxpooling is applied to the output generated by CNN/LSTM/BiLSTM layer to take the maximum score of each pattern. The main function of the pooling layer is to reduce the dimensionality of the CNN/LSTM/BiLSTM representations by down-sampling the output and keeping the maximum value. For reducing over-fitting by preventing complex co-adaptations on training data, a Dropout layer with a probability equal to 0.5 is added. The obtained scores are then feeded to a single feed-forward (fully-connected) layer with Relu activation. Finally, the output of that layer goes through a sigmoid layer that predicts the output classes. For all the models we used Adam optimisers with epoch 100 and an early\_stopping parameter for stopping the iteration in the absence of improvements. 

\section{Experimentation and Results}
\subsection{Experimental results}
For showing the impact of balanced/unbalanced corpus, we present the different results related to the detection of Hateful/non hateful detection separately. Table \ref{resultcorpus1} presents the results obtained on \textit{Corpus\_1}. It can be seen from Table \ref{resultcorpus1} that the F1-score obtained on the unbalanced corpus (\textit{Corpus\_1}) is up to 86\%. This result is obtained using the SG model associated to the CNN algorithm. Concerning the word2vec model, it can be seen that RF and LSVC algorithms give the best results for both SG and CBOW. However, deep learning classifiers (CNN, Bi-LSTM) associated to SG model of fastText outperformed the others classifiers. In addition, SG models outperforms CBOW models for all the used classifiers. 
\begin{table*}[h]
\begin{center}
\caption{Classification results on Corpus\_1}  
    \begin{tabular}{ccccccccccccc}
Corpus&Models&Type&ML Alg&\multicolumn{3}{c}{Hateful}&\multicolumn{3}{c}{Non-hateful}&\multicolumn{3}{c}{Average} \\
&&&& P&R&F1& P&R&F1& P&R&F1\\\hline
&&&GNB&0.32&0.80&0.46&0.91&0.56&0.69&0.79&0.61&0.64\\
&&&LR&0.70&0.19&0.30&0.82&0.98&0.89&0.80&0.81&0.77\\
&&SG&RF&0.69&0.33&0.44&0.84&0.96&0.90&0.81&0.83&0.80\\
&&&SGD&0.81&0.13&0.23&0.81&0.99&0.89&0.81&0.81&0.75\\
&&&LSVC&0.70&0.41&0.52&0.86&0.95&0.90&0.83&0.83&0.82\\ 
&Word2vec&&&&&&&&&&&\\
&&&GNB&0.30&0.82&0.44&0.91&0.48&0.63&0.78&0.55&0.59\\
&&&LR&0.75&0.04&0.07&0.79&1.00&0.88&0.79&0.79&0.71\\
&&CBOW&RF&0.50&0.17&0.26&0.81&0.95&0.88&0.75&0.79&0.75\\
&&&SGD&0.67&0.04&0.07&0.79&0.99&0.88&0.77&0.79&0.71\\
Corpus\_1&&&LSVC&0.57&0.15&0.24&0.81&0.97&0.88&0.76&0.80&0.75\\ 
\\
 
 &&&CNN&0.77&0.56&\textbf{0.65}&0.89&0.96&\textbf{0.92}&0.87&0.87&\textbf{0.86}\\
&&SG&LSTM&0.82&0.45&0.58&0.87&0.97&0.92&0.86&0.86&0.85\\
&&&Bi-LSTM&0.89&0.36&0.51&0.85&0.99&0.91&0.86&0.86&0.83\\
&FasText&&&&&&&&&&&\\
&&&CNN&0.71&0.46&0.56&0.87&0.95&0.91&0.84&0.85&0.83\\
&&CBOW&LSTM&0.67&0.53&0.59&0.88&0.93&0.90&0.83&0.84&0.84\\
&&&Bi-LSTM&0.56&0.61&0.59&0.89&0.87&0.88&0.82&0.82&0.82\\
\hline
\end{tabular}
  
  \label{resultcorpus1}
  \end{center}
\end{table*}

Table \ref{resultcorpus2} presents the results obtained on \textit{Corpus\_2}. It can be seen from Table \ref{resultcorpus2} that the F1-score obtained on the balanced corpus (\textit{Corpus\_2}) is up to 85\%. For the experiments using word2vec, LSVC seems to be the best choice. LSVC gives the best F1-score with  SG and CBOW (up to 0.80 with SG and up to 0.77 with CBOW). LR slightly outperforms LSVC on SG model (where F1-score is up to 0.81), however, the results decrease with CBOW model (up to 0.72). As well as to the results presented in Table \ref{resultcorpus1}, deep learning classifiers associated to SG model of fastText outperforms the others classifiers. The SG model also outperforms CBOW model for all the used classifiers, on this corpus as well.

From Both Tables \ref{resultcorpus1} and \ref{resultcorpus2} , it can be seen that the best F1-score obtained on \textit{Corpus\_1} (up to 86\%) is slightly better the F1-score obtained on \textit{Corpus\_1}. However only 65\% of hateful comment were correctly classifier using \textit{Corpus\_1}, where 83\% are correctly classified using \textit{Corpus\_2}. It also can be observed that deep learning classifiers are more appropriate with unbalanced data (F1-score up to 65\%) where the classic classifiers (GNB, LR, ect) are able to correctly classify only 52\%.
\begin{table*}[h]
\begin{center}
\caption{Classification results on Corpus\_2}  
    \begin{tabular}{ccccccccccccc}
Corpus&Models&Type&ML Alg&\multicolumn{3}{c}{Hateful}&\multicolumn{3}{c}{Non-hateful}&\multicolumn{3}{c}{Average} \\
&&&& P&R&F1& P&R&F1& P&R&F1\\\hline
 &&&GNB&0.63&0.82&0.71&0.83&0.63&0.71&0.74&0.71&0.71\\
&&&LR&0.79&0.75&0.77&0.82&0.85&0.84&0.81&0.81&0.81\\
&&SG&RF&0.81&0.62&0.71&0.76&0.89&0.82&0.78&0.78&0.77\\
&&&SGD&0.72&0.85&0.78&0.87&0.75&0.81&0.81&0.79&0.79\\
&&&LSVC&0.79&0.74&0.76&0.81&0.85&0.83&0.80&0.80&0.80\\ 
&Word2vec&&&&&&&&&&&\\
&&&GNB&0.54&0.85&0.66&0.80&0.45&0.58&0.69&0.62&0.61\\
&&&LR&0.72&0.58&0.65&0.73&0.83&0.77&0.72&0.72&0.72\\
&&CBOW&RF&0.73&0.63&0.68&0.75&0.82&0.78&0.74&0.74&0.74\\
&&&SGD&0.77&0.57&0.65&0.73&0.87&0.79&0.74&0.74&0.73\\
Corpus\_2&&&LSVC&0.75&0.70&0.72&0.79&0.82&0.80&0.77&0.77&0.77\\ 
\\
&&&CNN&0.86&0.69&0.77&0.80&0.92&0.85&0.83&0.82&0.82\\
&&SG&LSTM&0.93&0.60&0.73&0.76&0.97&0.85&0.83&0.81&0.80\\
&&&Bi-LSTM&0.85&0.81&\textbf{0.83}&0.86&0.89&\textbf{0.88}&0.85&0.86&\textbf{0.85}\\
&FasText&&&&&&&&&&&\\
 
&&&CNN&0.81&0.62&0.70&0.76&0.89&0.82&0.78&0.77&0.77\\
&&CBOW&LSTM&0.94&0.57&0.71&0.75&0.97&0.85&0.83&0.80&0.79\\
&&&Bi-LSTM&0.73&0.82&0.77&0.85&0.77&0.81&0.80&0.79&0.79\\ 
\hline
\end{tabular}
  
  \label{resultcorpus2}
  \end{center}
\end{table*}

\subsection{Perspective of improvements}
 The presented results are pretty good but they could be improved by integrating some pre-treatments. The first one is related to Arabizi transliteration. As Arabic people used both scripts Arabic and Arabizi. Handling them together or classifying Arabizi without calling the transliteration step could give wrong results. We previously showed that the transliteration consequently improved the results of sentiment analysis \cite{guellil2018arabizi}. 
 We previously present a transliteration based on rules-based approach \cite{guellil2018arabizi,guellil2018approche} but we conclude that a corpus based approach would certainly improve the results. Hence, we plan to propose a corpus-based approach for transliteration and apply this approach on the annotated corpus for having one script used for Arabic language. In addition to scripts, Arabic people also use other languages to express their opinions in social media, such as French or English. However, the proportion of these language is not really important comparing to the proportion of Arabic and Arabizi. In the context of this study, we handle all the languages in the same corpus. However, a language identification step would consequently improve the results. Hence, as an improvement to this work, we plan to propose an identification approach between Arabizi, French and English (because they share the same script). 
 
 Also, this study is the first one, to the best of our knowledge presenting a publicly available corpus\footnote{Freely available for the research purpose, after the paper acceptance} dedicated to Arabic hate-speech detection against women. This resource was manually annotated by three native annotators following a guideline and standards. We are actually working on extending this resource. We are first targeting 10,000 comments. We are also working on automatic techniques in order to increase this corpus. Our principal aim is to relate our previous studies \cite{guellil2018sentialg,guellil2018arabizi,imane2019set} related to sentiment analysis to the hate speech detection.
\section{Conclusion}
Hate speech detection is a research area attracting the research community interest more and more. Different studies have been proposed and most of them are quietly recent (during 2016 and 2019). The purpose of this studies is mitigated between the detection of hate speech in general and and hate speech targeting a special community or a special group. In this context, the principal aim of this paper is to detect hate speech against women in Arabic community on social media. We automatically collected data related to women from Youtube. Afterwards, we randomly select 5,000 comments and give them to three annotators in order to labelled them as hateful or non-hateful. However, for increasing the precision, we concentrate on the portion of the corpus were all the annotators were agree. It allow us to construct a corpus containing 3,798 comments (where 3.006 are non-hateful and 792 are hateful). We also constructed a balanced corpus containing 1,798 comment randomly picked up from the aforementioned one. For validating the constructed corpus, we used different machine learning algorithm such as LSVC, GNB, SGD, etc and deep learning one such as CNN? LSTM, etc. However, The exeperimental results showed that the deep learning classifiers (especially CNN, Bi-LSTM) outperform the other classifiers by respectively achieving and F1-score up to 86\%.

For improving this work we plan to integrate a transliteration system for transforming Arabizi to Arabic. We also plan to identify the different language before proceeding to the classification. Finally, we also plan to automatically increase the training corpus.

\bibliographystyle{spmpsci}
\bibliography{sample-bibliography}

\end{document}